\documentclass[11pt,a4paper]{article}
\usepackage[hyperref]{emnlp2020}
\usepackage{times}
\usepackage{latexsym}

\usepackage{microtype}

\usepackage[absolute,overlay]{textpos}

\usepackage{arydshln}
\usepackage{graphicx}
\usepackage{booktabs}
\usepackage{tabularx}
\usepackage{amssymb}  % for checkmark
\usepackage{subcaption}
\usepackage{amsmath} % for aligning equations

\aclfinalcopy % Uncomment this line for the final submission
\usepackage{listings} % python code
\usepackage[utf8]{inputenc}
\usepackage{xcolor}

\definecolor{codegreen}{rgb}{0,0.6,0}
\definecolor{codegray}{rgb}{0.5,0.5,0.5}
\definecolor{codepurple}{rgb}{0.58,0,0.82}
\definecolor{backcolour}{rgb}{0.95,0.95,0.92}

\lstdefinestyle{mystyle}{
    backgroundcolor=\color{backcolour},   
    commentstyle=\color{codegreen},
    keywordstyle=\color{magenta},
    numberstyle=\tiny\color{codegray},
    stringstyle=\color{codepurple},
    basicstyle=\ttfamily\footnotesize,
    breakatwhitespace=false,         
    breaklines=true,                 
    captionpos=b,                    
    keepspaces=true,                 
    numbers=left,                    
    numbersep=5pt,                  
    showspaces=false,                
    showstringspaces=false,
    showtabs=false,                  
    tabsize=2
}

\usepackage{pifont} % \ding
\newcommand{\one}{\ding{172}\hspace{0.3mm}}
\newcommand{\two}{\ding{173}\hspace{0.3mm}}
\newcommand{\three}{\ding{174}\hspace{0.3mm}}
\newcommand{\four}{\ding{175}\hspace{0.3mm}}
\newcommand{\five}{\ding{176}\hspace{0.3mm}}
\newcommand{\six}{\ding{177}\hspace{0.3mm}}

\title{AdapterHub: A Framework for Adapting Transformers}

\author{ Jonas Pfeiffer\thanks{*Equal contribution.} $^{ 1}$, Andreas R\"uckl\'{e}$^{*1}$, Clifton Poth$^{*1}$, \\{\bf Aishwarya Kamath$^{2}$,  Ivan Vuli\'{c}$^{4}$, Sebastian Ruder$^{5}$,} \\ {\bf Kyunghyun Cho$^{2,3}$, Iryna Gurevych$^{1}$} \\
$^1$%Ubiquitous Knowledge Processing Lab, 
  Technical University of Darmstadt \\
  $^2$New York University \hspace{0.5em}  $^3$CIFAR Associate Fellow\\
    $^4$%Language Technology Lab, 
    University of Cambridge \hspace{0.5em} \\
$^5$DeepMind \\
\texttt{\href{https://AdapterHub.ml}{AdapterHub.ml}} \\}

\date{}

\begin{document}
\maketitle
\begin{abstract}
The current modus operandi in NLP involves downloading and fine-tuning pre-trained models consisting of hundreds of millions, or even billions of parameters. Storing and sharing such large trained models is expensive, slow, and time-consuming, which impedes progress towards more general and versatile NLP methods that learn from and for many tasks. Adapters---small learnt bottleneck layers inserted within each layer of a pre-trained model--- ameliorate this issue by avoiding full fine-tuning of the entire model. However, sharing and integrating adapter layers is not straightforward. We propose AdapterHub, a framework that allows dynamic ``stiching-in" of pre-trained adapters for different tasks and languages. The framework, built on top of the popular HuggingFace Transformers library, enables extremely easy and quick adaptations of state-of-the-art pre-trained models (e.g., BERT, RoBERTa, XLM-R) across tasks and languages. Downloading, sharing, and training adapters is as seamless as possible using minimal changes to the training scripts and a specialized infrastructure. Our framework enables scalable and easy access to sharing of task-specific models, particularly in low-resource scenarios. AdapterHub includes all recent adapter architectures and can be found at \href{https://AdapterHub.ml}{AdapterHub.ml}.
\end{abstract}

\begin{textblock*}{1cm}(15cm,3.3cm) % {block width} (coords)
\includegraphics[width=3.8cm]{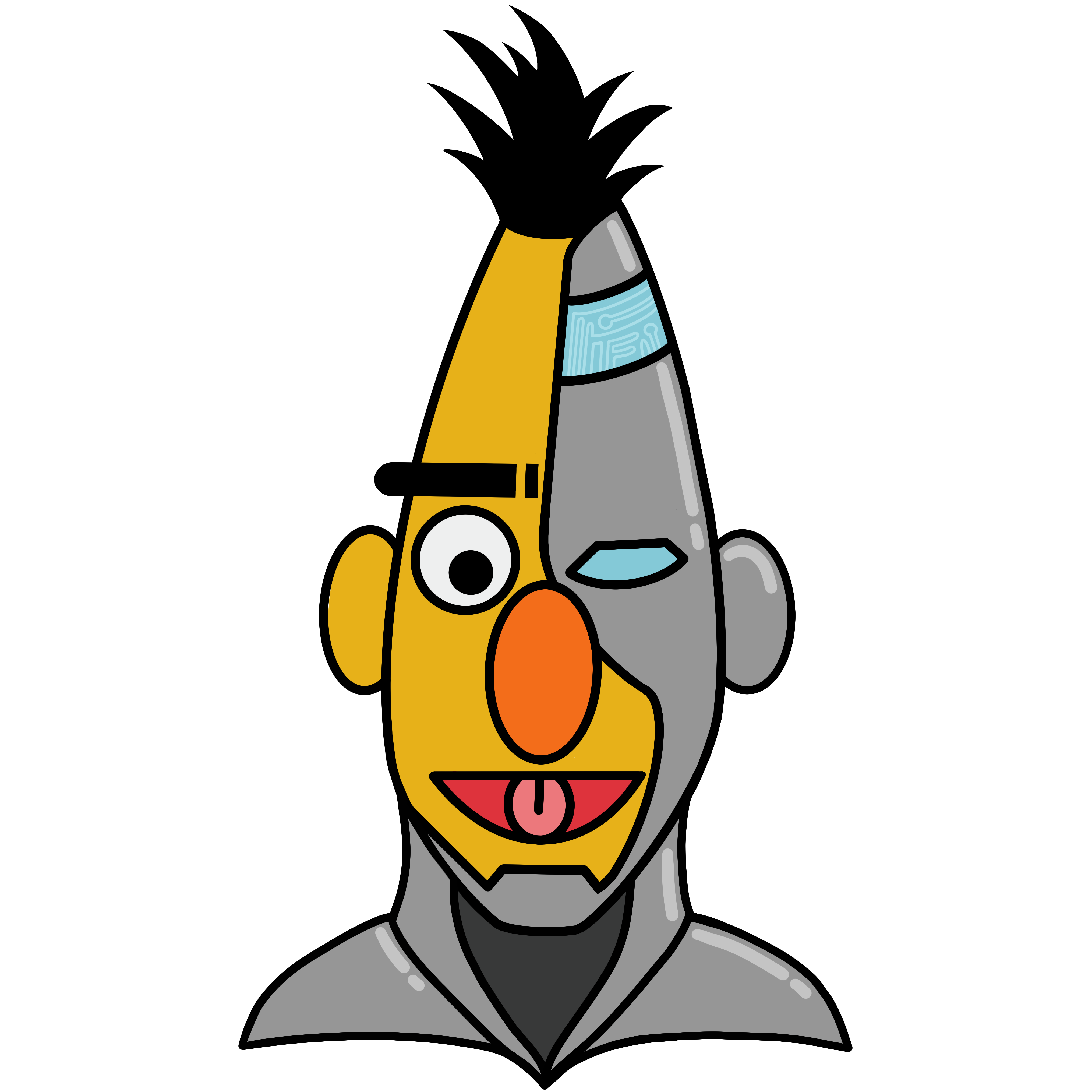}
\end{textblock*}
 
 \section{Introduction}
 
Recent advances in NLP leverage transformer-based language models \cite{Vaswani2017transformer}, pre-trained on large amounts of text data \cite{Devlin2019bert, liu2019roberta, Conneau2020xlm-r}. These models are fine-tuned on a target task and achieve state-of-the-art (SotA) performance for most natural language understanding tasks. Their performance has been shown to scale with their size \cite{Kaplan2020scalinglaws} and recent models have reached billions of parameters \cite{Raffel2019t3,Brown:2020gpt3}. While fine-tuning large pre-trained models on target task data can be done fairly efficiently \cite{Howard2018ulmfit}, training them for multiple tasks and sharing trained models is often prohibitive. This precludes research on more modular architectures \cite{Shazeer2017mixtureofexperts}, task composition \cite{Andreas2016learningtocompose}, and injecting biases and external information (e.g., world or linguistic knowledge) into large models \cite{Lauscher:2019libert,Wang20kadapters}.

\textit{Adapters} \cite{Houlsby2019adapters} have been introduced as an alternative lightweight fine-tuning strategy that achieves on-par performance to full fine-tuning \cite{peters2019tune} on most tasks. They consist of a small set of additional newly initialized weights at every layer of the transformer. These weights are then trained during fine-tuning, while the pre-trained parameters of the large model are kept frozen/fixed. This enables efficient parameter sharing between tasks by training many task-specific and language-specific adapters for the same model, which can be exchanged and combined post-hoc. Adapters have recently achieved strong results in multi-task and cross-lingual transfer learning \cite{Pfeiffer2020adapterfusion,pfeiffer20madx}.

However, reusing and sharing adapters is not straightforward. Adapters are rarely released individually; their architectures differ in subtle yet important ways, and they are model, task, and language dependent. To mitigate these issues and facilitate transfer learning with adapters in a range of settings, we propose AdapterHub, a framework that enables seamless training and sharing of adapters.

AdapterHub is built on top of the popular \texttt{transformers} framework by HuggingFace\footnote{\href{https://github.com/huggingface/transformers}{https://github.com/huggingface/transformers}} \cite{Wolf2019transformers}, which provides access to state-of-the-art pre-trained language models. We enhance \texttt{transformers} with adapter modules that can be combined with existing SotA models with minimal code edits. We additionally provide a website that enables quick and seamless upload, download, and sharing of pre-trained adapters. AdapterHub is available online at: \href{https://AdapterHub.ml}{AdapterHub.ml}.

AdapterHub for the first time enables NLP researchers and practitioners to easily and efficiently share and obtain access to models that have been trained for particular tasks, domains, and languages. This opens up the possibility of building on and combining information from many more sources than was previously possible, and makes research such as intermediate task training \cite{Pruksachatkun2020}, composing information from many tasks \cite{Pfeiffer2020adapterfusion}, and training models for very low-resource languages \cite{pfeiffer20madx} much more accessible.

\vspace{1.5mm}
\noindent \textbf{Contributions.} 1) We propose an easy-to-use and extensible adapter training and sharing framework for transformer-based models such as BERT, RoBERTa, and XLM(-R); 2) we incorporate it into the HuggingFace \texttt{transformers} framework, requiring as little as two additional lines of code to train adapters with existing scripts; 3) our framework automatically extracts the adapter weights, storing them separately to the pre-trained transformer model, requiring as little as  1Mb of storage; 4) we provide an open-source framework and website that allows the community to upload their adapter weights, making them easily accessible with only one additional line of code; 5) we incorporate adapter composition  as well as adapter stacking out-of-the-box and pave the way for a wide range of other extensions in the future.

\section{Adapters}
\label{sec:Adapters}
While  the predominant methodology for transfer learning is to fine-tune all weights of the pre-trained model, \emph{adapters} have recently been introduced as an alternative approach, with applications in computer vision \cite{rebuffi2017learning} as well as the NLP domain \cite{Houlsby2019adapters, Bapna2019adapters, Wang20kadapters,Pfeiffer2020adapterfusion, pfeiffer20madx}.

\subsection{Adapter Architecture}
\label{sec:Adapter_Architectur}

Adapters are neural modules with a small amount of additional newly introduced parameters  $\Phi$ within a large pre-trained model with parameters $\Theta$. The parameters $\Phi$ are learnt on a target task while keeping $\Theta$ fixed; $\Phi$ thus learn to encode task-specific representations in intermediate layers of the pre-trained model. 
Current work predominantly focuses on training adapters for each task separately  \cite{Houlsby2019adapters,Bapna2019adapters,Pfeiffer2020adapterfusion,pfeiffer20madx}, which enables parallel training and subsequent combination of the weights.

In NLP, adapters have been mainly used within deep transformer-based architectures \cite{Vaswani2017transformer}. At each transformer layer $l$, a set of adapter parameters $\Phi_l$ is introduced. 
The placement and architecture of adapter parameters $\Phi$ within a pre-trained model is non-trivial and may impact their efficacy: \newcite{Houlsby2019adapters} experiment with different adapter architectures, empirically validating that a two-layer feed-forward neural network with a bottleneck works well. 
While this down- and up-projection has largely been agreed upon, the actual placement of adapters within each transformer block, as well as the introduction of new LayerNorms\footnote{Layer normalization learns to normalize the inputs across the features. This is usually done by introducing a new set of features for mean and variance.} \cite{Ba16_LayerNorm} varies in the literature \cite{Houlsby2019adapters,Bapna2019adapters, pmlr-v97-stickland19a, Pfeiffer2020adapterfusion}. In order to support standard adapter architectures from the literature, as well as to enable easy extensibility, AdapterHub provides a configuration file where the architecture settings can be defined dynamically. We illustrate the different configuration possibilities in Figure~\ref{fig:AdapterHub_config}, and describe them in more detail in \S\ref{sec:Technical_Repository}.

\subsection{Why Adapters?}
\label{sec:Adapter_Why_Adapters}

Adapters provide numerous benefits over fully fine-tuning a model such as scalability, modularity, and composition. We now provide a few use-cases for adapters to illustrate their usefulness in practice.

\vspace{1.5mm}
\noindent \textbf{Task-specific Layer-wise Representation Learning.}
Prior to the introduction of adapters, in order to achieve SotA performance on downstream tasks,  the entire pre-trained transformer model needs to be fine-tuned \cite{peters2019tune}. Adapters have been shown to work on-par with full fine-tuning, by adapting the representations at every layer. We present the results of fully fine-tuning the model compared to two different adapter architectures on the GLUE benchmark \cite{wang18glue} in Table~\ref{table:scores}. The adapters of \citet[][Figure \ref{fig:arch_houlsby}]{Houlsby2019adapters} and \citet[][Figure \ref{fig:arch_pfeiffer}]{Pfeiffer2020adapterfusion}  comprise two and one down- and up-projection within each transformer layer, respectively. The former adapter thus has more capacity at the cost of training and inference speed. We find that for all settings, there is no large difference in terms of performance between the model architectures, verifying that training adapters is a suitable and lightweight alternative to full fine-tuning in order to achieve SotA performance on downstream tasks.

\begin{table}[!t]
\centering
{\small
\begin{tabularx}{\linewidth}{lXXX}
\toprule
      & \textbf{Full} & \textbf{Pfeif.} & \textbf{Houl.} \\
      \midrule
\textbf{RTE} \cite{wang18glue}   &   66.2   &     70.8     &    69.8     \\
\textbf{MRPC} \cite{dolan2005automatically}  &   90.5   &     89.7     &    91.5     \\ 
\textbf{STS-B} \cite{cer2017semeval} &   88.8   &     89.0     &    89.2     \\
\textbf{CoLA} \cite{warstadt2018neural}  &   59.5   &     58.9     &    59.1     \\
\textbf{SST-2} \cite{socher2013-sst} &   92.6   &     92.2     &    92.8     \\
\textbf{QNLI} \cite{rajpurkar2016squad}  &   91.3   &     91.3     &    91.2     \\
\textbf{MNLI} \cite{Williams18MNLI}  &   84.1   &     84.1     &    84.1     \\
\textbf{QQP} \cite{WinNT}   &   91.4   &     90.5     &    90.8     \\ 
\bottomrule
\end{tabularx}}%
\caption{Mean development scores over 3 runs on GLUE \cite{wang18glue} leveraging the BERT-Base pre-trained weights. We present the results with full fine-tuning (\textbf{Full}) and with the adapter architectures of \citet[][\textbf{Pfeif.}, Figure \ref{fig:arch_pfeiffer}]{Pfeiffer2020adapterfusion} and \citet[][\textbf{Houl.}, Figure \ref{fig:arch_houlsby}]{Houlsby2019adapters} both with bottleneck size 48. We show F1 for MRPC, Spearman rank correlation for STS-B, and accuracy for the rest. RTE is a combination of datasets \cite{dagan2005pascal, bar2006second, giampiccolo2007third}. }
\label{table:scores}
\end{table}

\vspace{1.5mm}
\noindent \textbf{Small, Scalable, Shareable.}
Transformer-based models are very deep neural networks with millions or billions of weights and large storage requirements, e.g., around 2.2Gb of compressed storage space is needed for XLM-R Large \cite{Conneau2020xlm-r}.  Fully fine-tuning these models for each task separately requires storing a copy of the fine-tuned model for each task. This impedes both iterating and parallelizing training, particularly in storage-restricted environments.

Adapters mitigate this problem. Depending on the model size and the adapter bottleneck size, a single task requires as little as 0.9Mb storage space. We present the storage requirements in Table~\ref{table:paramsizes}. This highlights that $>99\%$ of the parameters required for each target task are fixed during training and can be shared across all models for inference. For instance, for the popular Bert-Base model with a size of 440Mb, storing 2 fully fine-tuned models amounts to the same storage space required by 125 models with adapters, when using a bottleneck size of 48 and adapters of \citet{Pfeiffer2020adapterfusion}. Moreover, when performing inference on a mobile device, adapters can be leveraged to save a significant amount of storage space, while supporting a large number of target tasks. Additionally, due to the small size of the adapter modules---which in many cases do not exceed the file size of an image---new tasks can be added on-the-fly. 
Overall, these factors make adapters a much more computationally---and ecologically \cite{StrubellGM19}---viable option compared to updating entire models \cite{rueckle2020adapterdrop}. Easy access to fine-tuned models may also improve reproducibility as researchers will be able to easily rerun and evaluate trained models of previous work. 

\begin{table}[t]
\centering
{\small
\begin{tabularx}{\linewidth}{cXXXX}
\toprule
   & \multicolumn{2}{X}{\textbf{Base}} & \multicolumn{2}{X}{\textbf{Large}} \\
   CRate & \#Params      & Size     & \#Params      & Size      \\
   \midrule
\textbf{64}  &     0.2M          &   0.9Mb   &   0.8M      &    3.2Mb       \\
\textbf{16} &     0.9M      &   3.5Mb  &     3.1M     &      13Mb     \\
\textbf{2} &     7.1M     &  28Mb   &     25.2M     &   97Mb       \\
\bottomrule
\end{tabularx}}%
\caption{Number of additional parameters and compressed storage space of the adapter of \citet{Pfeiffer2020adapterfusion} in (Ro)BERT(a)-Base and Large transformer architectures. The adapter of \citet{Houlsby2019adapters} requires roughly twice as much space. 
\textit{CRate} refers to the adapter's compression rate: e.g., a. rate of 64 means that the adapter's bottleneck layer is 64 times smaller than the underlying model's hidden layer size.}
\label{table:paramsizes}
\end{table}

\vspace{1.5mm}
\noindent \textbf{Modularity of Representations.}
Adapters learn to encode information of a task within designated parameters. Due to the encapsulated placement of adapters, wherein the surrounding parameters are fixed, at each layer an adapter is forced to learn an output representation compatible with the subsequent layer of the transformer model. This setting allows for modularity of components such that adapters can be stacked on top of each other, or replaced dynamically. In a recent example, \citet{pfeiffer20madx} successfully combine adapters that have been independently trained for specific tasks and languages. This demonstrates that adapters are modular and that output representations of different adapters are compatible. As NLP tasks become more complex and require knowledge that is not directly accessible in a single monolithic pre-trained model \cite{ruder2019transfer}, adapters will provide NLP researchers and practitioners with many more sources of relevant information that can be easily combined in an efficient and modular way.

\vspace{1.5mm}
\noindent \textbf{Non-Interfering Composition of Information.} Sharing information across tasks has a long-standing history in machine learning \cite{Ruder2017}.  
Multi-task learning (MTL), which shares a set of parameters between tasks, has arguably received the most attention. However, MTL suffers from problems such as catastrophic forgetting where information learned during earlier stages of training is ``overwritten'' \cite{Masson2019episodicmemory}, catastrophic interference where the performance of a set of tasks deteriorates when adding new tasks \cite{Hashimoto2017ajointmanytask}, and intricate task weighting for tasks with different distributions \cite{Sanh2019hierarchical}. 

The encapsulation of adapters forces them to learn output representations that are compatible across tasks. When training adapters on different downstream tasks, they store the respective information in their designated parameters. Multiple adapters can then be combined, e.g., with attention \cite{Pfeiffer2020adapterfusion}. Because the respective adapters are trained separately, the necessity of sampling heuristics due to skewed data set sizes no longer arises. By separating knowledge extraction and composition, adapters mitigate the two most common pitfalls of multi-task learning, catastrophic forgetting and catastrophic interference.

Overcoming these problems together with the availability of readily available trained task-specific adapters enables researchers and practitioners to leverage information from specific tasks, domains, or languages that is often more relevant for a specific application---rather than more general pre-trained counterparts. Recent work \cite{Howard2018ulmfit,phang2018sentence,Pruksachatkun2020,gururangan2020don} has shown the benefits of such information, which was previously only available by fully fine-tuning a model on the data of interest prior to task-specific fine-tuning.

\section{AdapterHub}

\begin{figure}[!t] %[htp]
\centering
\includegraphics[width=0.94\linewidth]{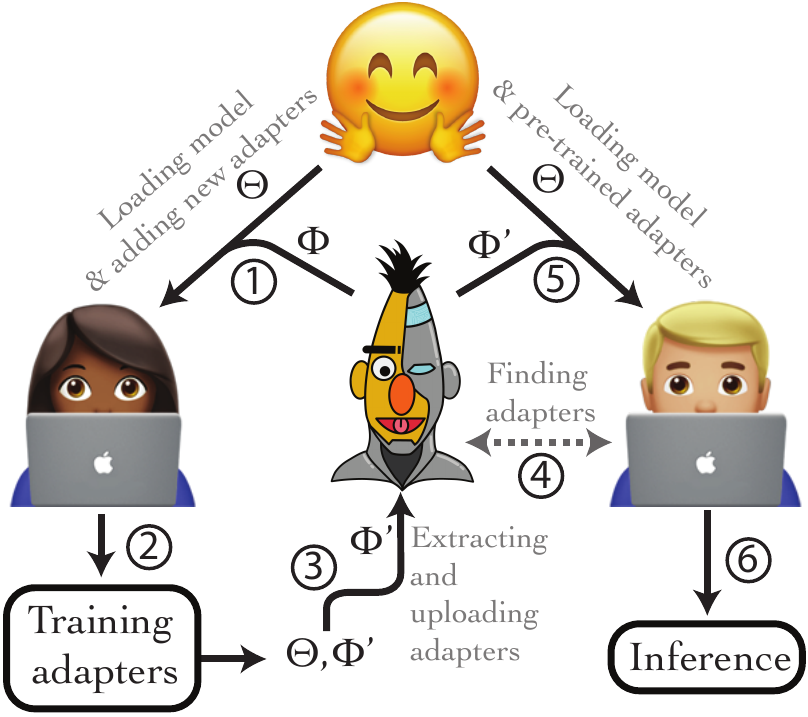}
\caption{The AdapterHub Process graph. Adapters $\Phi$ are introduced into a pre-trained transformer $\Theta$ (step \one) and are trained (\two). They can then be extracted and open-sourced (\three) and visualized (\four). Pre-trained adapters are downloaded on-the-fly (\five) and stitched into a model that is used for inference (\six).}
\label{fig:Process} 
\end{figure}

AdapterHub consists of two core components: \textbf{1)} A library built on top of HuggingFace \texttt{transformers}, and \textbf{2)} a website that dynamically provides analysis and filtering of pre-trained adapters. AdapterHub provides tools for the entire life-cycle of adapters, illustrated in Figure~\ref{fig:Process} and discussed in what follows: \one~introducing new adapter weights $\Phi$ into pre-trained transformer weights $\Theta$;  \two~training adapter weights $\Phi$ on a downstream task (while keeping $\Theta$ frozen); \three~automatic extraction of the trained adapter weights $\Phi'$ and open-sourcing the adapters; \four~automatic visualization of the adapters with configuration filters; \five~on-the-fly downloading/caching the pre-trained adapter weights $\Phi'$ and stitching the adapter into the pre-trained transformer model $\Theta$; \six~performing inference with the trained adapter transformer model.

\begin{figure*}[htp]
\centering
\begin{minipage}{0.999\linewidth}
\begin{lstlisting}[language=Python]
from transformers import AutoModelForSequenceClassification, AdapterType
model = AutoModelForSequenceClassification.from_pretrained("roberta-base")
model.add_adapter("sst-2", AdapterType.text_task, config="pfeiffer")
model.train_adapter(["sst-2"])
# Train model ...
model.save_adapter("adapters/text-task/sst-2/", "sst-2")
# Push link to zip file to AdapterHub ...
\end{lstlisting}
\end{minipage} 
\caption{\one~Adding new adapter weights $\Phi$ to pre-trained RoBERTa-Base weights $\Theta$ (line 3), and freezing $\Theta$  (line 4). \three~Extracting and storing the trained adapter weights $\Phi'$ (line 6). }
\label{fig:code_add_adapter} 
\end{figure*}

\subsection*{\one~Adapters in Transformer Layers} 
\label{sec:Technical_Repository}

We minimize the required changes to existing HuggingFace training scripts, resulting in only two additional lines of code. In Figure~\ref{fig:code_add_adapter} we present the required code to add adapter weights (line 3) and freeze all the transformer weights $\Theta$ (line 4). In this example, the model is prepared to train a task adapter on the binary version of the Stanford Sentiment Treebank \cite[SST;][]{socher2013-sst} using the adapter architecture of \citet{Pfeiffer2020adapterfusion}. Similarly, language adapters can be added by setting the type parameter to \texttt{AdapterType.text\_language}, and other adapter architectures can be chosen accordingly. 

\begin{figure}[t!]
\centering
\begin{subfigure}{.4775\linewidth}
  \centering
  \includegraphics[width=\linewidth]{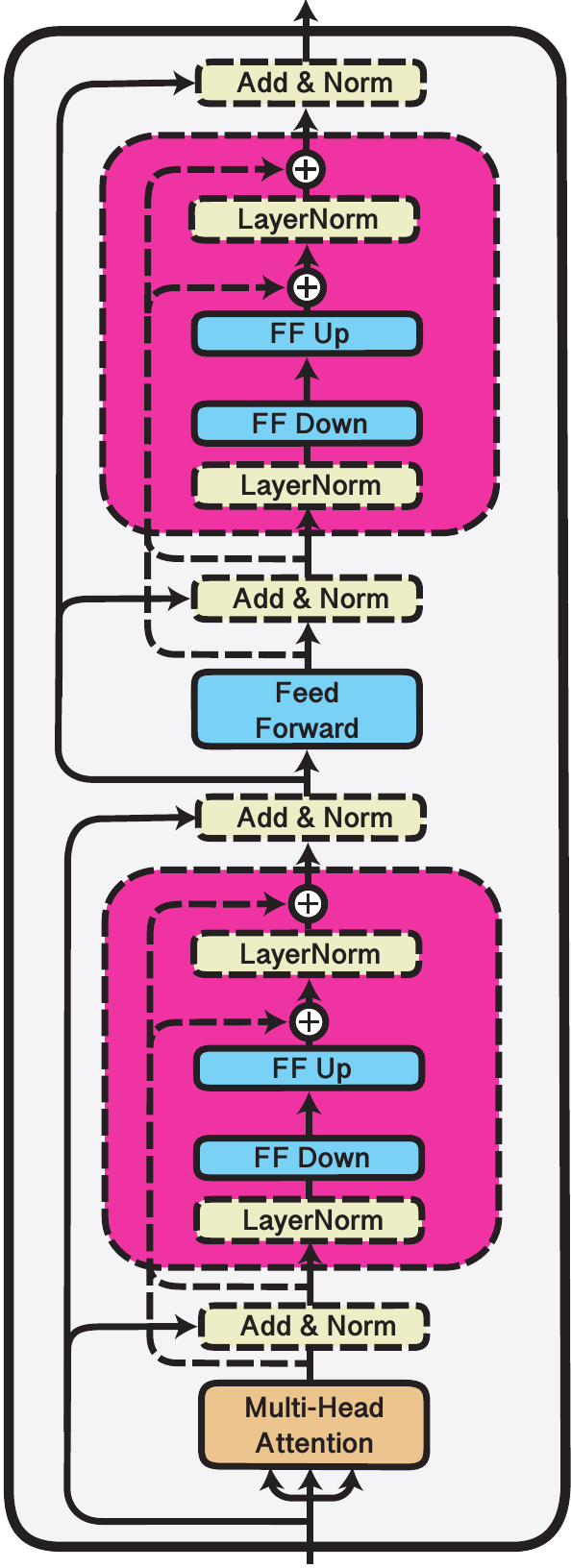}%\hfill
  \caption{Configuration Possibilities}
  \label{fig:config}
\end{subfigure}%
\hfill
\hfill
\begin{subfigure}{.4\linewidth}
    \begin{subfigure}{\linewidth}
      \centering
      \includegraphics[width=0.9\linewidth]{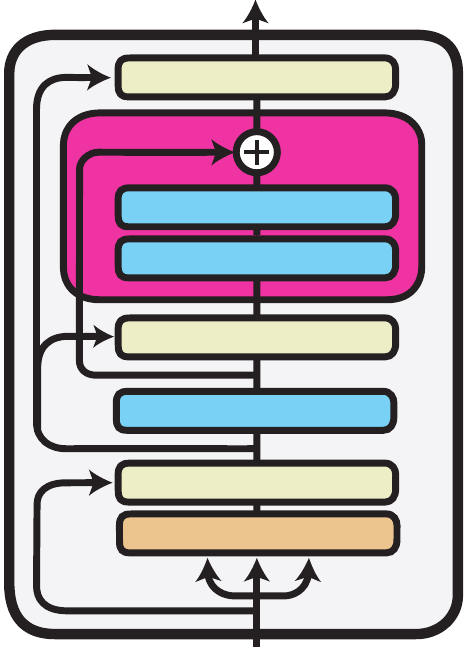}\par\medskip
      \caption{Pfeiffer Architecture}
      \label{fig:arch_pfeiffer}
    \end{subfigure}
    \par\medskip
        \begin{subfigure}{\linewidth}
      \centering
      \includegraphics[width=0.9\linewidth]{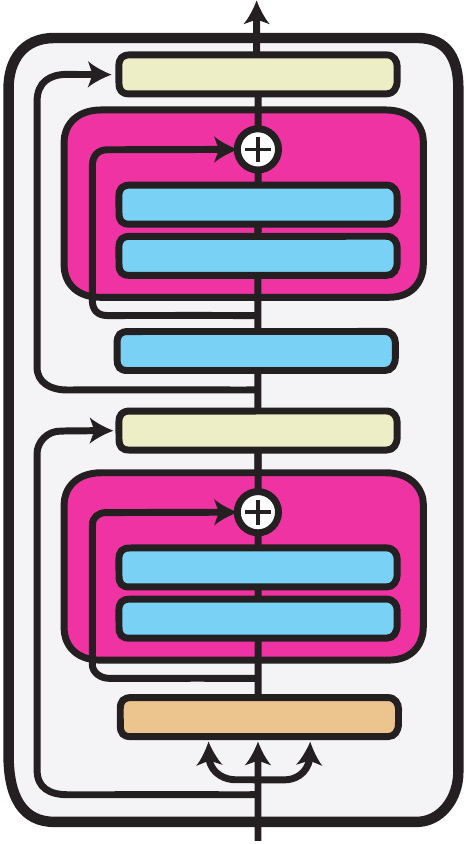}\par\medskip
      \caption{Houlsby Architecture}
      \label{fig:arch_houlsby}
    \end{subfigure}
\end{subfigure} 
\caption{Dynamic customization possibilities where dashed lines in (a) show the current configuration options. These options include the placements of new weights $\Phi$ (including down and up projections as well as new LayerNorms), residual connections, bottleneck sizes as well as activation functions. All new weights $\Phi$ are illustrated within the pink boxes, everything outside belongs to the pre-trained weights $\Theta$. In addition, we provide pre-set configuration files for architectures in the literature. The resulting configurations for the architecture proposed by \citet{Pfeiffer2020adapterfusion} and \citet{Houlsby2019adapters} are illustrated in (b) and (c) respectively. We also provide a configuration file for the architecture proposed by \newcite{Bapna2019adapters}, not shown here. }
\label{fig:AdapterHub_config} 
\end{figure}

While we provide ready-made configuration files for well-known architectures in the current literature, adapters are dynamically configurable, which makes it possible to define a multitude of architectures. We illustrate the configurable components as dashed lines and objects in Figure~\ref{fig:AdapterHub_config}. The configurable components are placements of new weights, residual connections as well as placements of LayerNorm layers \cite{Ba16_LayerNorm}. 

The code changes within the HuggingFace \texttt{transformers} framework are realized through \texttt{MixIns}, which are inherited by the respective transformer classes. This minimizes the amount of code changes of our proposed extensions and encapsulates adapters as designated classes. It further increases readability as adapters are clearly separated from the main \texttt{transformers} code base, which makes it easy to keep both repositories in sync as well as to extend AdapterHub.

\subsection*{\two~Training Adapters}

Adapters are trained in the same manner as full fine-tuning of the model. The information is passed through the different layers of the transformer where additionally to the pre-trained weights at every layer the representations are additionally passed through the adapter parameters. However, in contrast to full fine-tuning, the pre-trained weights $\Theta$ are fixed and only the adapter weights $\Phi$ and the prediction head are trained. Because $\Theta$ is fixed, the adapter weights $\Phi$ are encapsuled within the transformer weights, forcing them to learn compatible representations across tasks. 
\begin{figure*}[!th]
\centering
\begin{minipage}{0.975\linewidth}
\begin{lstlisting}[language=Python]
from transformers import AutoModelForSequenceClassification, AdapterType
model = AutoModelForSequenceClassification.from_pretrained("roberta-base")
model.load_adapter("sst-2", config="pfeiffer")
\end{lstlisting}
\end{minipage} 
\caption{\five~After the correct adapter has been identified by the user on the explore page of \href{https://AdapterHub.ml/explore/}{AdapterHub.ml}, they can load and stitch the pre-trained adapter weights $\Phi'$ into the transformer $\Theta$ (line 3).}
\label{fig:code_load_adapter} 
\end{figure*}

\subsection*{\three~Extracting and Open-Sourcing Adapters }
When training adapters instead of full fine-tuning, it is no longer necessary to store checkpoints of the entire model. Instead, only the adapter weights $\Phi'$, as well as the prediction head need to be stored, as the base model's weights $\Theta$ remain the same. 
This is integrated automatically as soon as adapters are trained, which significantly reduces the required storage space during training and enables storing a large number of checkpoints simultaneously.

When adapter training has completed, the parameter file together with the corresponding adapter configuration file are zipped and uploaded to a public server. The user then enters the metadata (e.g., URL to weights, user info, description of training procedure, data set used, adapter architecture, GitHub handle, Twitter handle) into a designated YAML file and issues a pull request to the AdapterHub GitHub repository. When all automatic checks pass, the \href{https://AdapterHub.ml}{AdapterHub.ml} website is automatically regenerated with the newly available adapter, which makes it possible for users to immediately find and use these new weights described by the metadata. We hope that the ease of sharing pre-trained adapters will further facilitate and speed up new developments in transfer learning in NLP.

\subsection*{\four~Finding Pre-Trained Adapters }

The website \href{https://adapterhub.ml}{AdapterHub.ml} provides a dynamic overview of the currently available pre-trained adapters. Due to the large number of tasks in many different languages as well as different transformer models, we provide an intuitively understandable hierarchical structure, as well as search options. This makes it easy for users to find adapters that are suitable for their use-case. Namely, AdapterHub's \href{https://adapterhub.ml/explore/}{explore} page is structured into three hierarchical levels. At the \textit{first} level, adapters can be viewed by task or language. The \textit{second} level allows for a more fine-grained distinction separating adapters into data sets of higher-level NLP tasks following a categorization similar to \href{https://paperswithcode.com/area/natural-language-processing}{paperswithcode.com}. 
For languages, the second level distinguishes the adapters by the language they were trained on. The \textit{third} level separates adapters into individual datasets or domains such as SST for sentiment analysis or Wikipedia for Swahili.

When a specific dataset has been selected, the user can see the available pre-trained adapters for this setting. Adapters depend on the transformer model they were trained on and are otherwise \textit{not} compatible.\footnote{We plan to look into mapping adapters between different models as future work.} The user selects the model architecture and certain hyper-parameters and is shown the compatible adapters. When selecting one of the adapters, the user is provided with additional information about the adapter, which is available in the metadata (see \three~again for more information).

\subsection*{\five~Stitching-In Pre-Trained Adapters }
Pre-trained adapters can be stitched into the large transformer model as easily as adding randomly initialized weights; this requires a single line of code, see Figure~\ref{fig:code_load_adapter}, line 3. When selecting an adapter on the website (see \four ~again) the user is provided with sample code, which corresponds to the configuration necessary to include the specific weights.\footnote{When selecting an adapter based on a name, we allow for string matching as long as there is no ambiguity.}

\subsection*{\six~Inference with Adapters }

Inference with a pre-trained model that relies on adapters is in line with the standard inference practice based on full fine-tuning. Similar to \textit{training} adapters, during inference the active adapter name is passed into the model together with the text tokens. At every transformer layer the information is passed through the transformer layers and the corresponding adapter parameters. 

The adapters can be used for inference in the designated task they were trained on. To this end, we provide an option to upload the prediction heads together with the adapter weights. In addition, they can be used for further research such as transferring the adapter to a new task, stacking multiple adapters, fusing the information from diverse adapters, or enriching AdapterHub with adapters for other modalities, among many other possible modes of usage and future directions.

\section{Conclusion and Future Work}
\label{sec:Conclusion}

We have introduced AdapterHub, a novel easy-to-use framework that enables simple and effective transfer learning via training and community sharing of \textit{adapters}. Adapters are small neural modules that can be stitched into large pre-trained transformer models to facilitate, simplify, and speed up transfer learning across a range of languages and tasks. AdapterHub is built on top of the commonly used HuggingFace \texttt{transformers}, and it requires only adding as little as two lines of code to existing training scripts. 
Using adapters in AdapterHub has numerous benefits such as improved reproducibility, much better efficiency compared to full fine-tuning, easy extensibility to new models and new tasks, and easy access to trained models.

With AdapterHub, we hope to provide a suitable and stable framework for the community to train, search, and use adapters. We plan to continuously improve the framework, extend the composition and modularity possibilities, and support other transformer models, even the ones yet to come.

\section*{Acknowledgments}
Jonas Pfeiffer is supported by the LOEWE initiative (Hesse, Germany) within the emergenCITY center. 
Andreas R\"uckl\'{e} is supported by the German Federal Ministry of Education and Research and the Hessen State Ministry for Higher Education, Research and the Arts within their joint support of the National Research Center for Applied Cybersecurity ATHENE, and by the German Research Foundation under grant EC 503/1-1 and GU 798/21-1. Aishwarya Kamath  is supported in part by a DeepMind PhD Fellowship. The work of Ivan Vuli\'{c} is supported by the ERC Consolidator Grant LEXICAL: Lexical Acquisition Across Languages (no 648909). Kyunghyun Cho is supported by  Samsung Advanced Institute of Technology (Next Generation Deep Learning: from pattern recognition to AI) and Samsung Research (Improving Deep Learning using Latent Structure).

 We would like to thank \href{https://instagram.com/isabelpfeiffer_art?igshid=165k7u2wz7pb0}{Isabel Pfeiffer} for the illustrations.

\bibliography{emnlp2020}
\bibliographystyle{acl_natbib}

\end{document}